\documentclass[a4paper, 10pt, conference]{ieeeconf}      

\IEEEoverridecommandlockouts                              %
\usepackage{graphics} 
\usepackage{epsfig} 
\usepackage{mathptmx} 
\usepackage{times} 
\usepackage{amsmath} 
\usepackage{amssymb}  
\usepackage{url}

\begin{document}

\title{\LARGE \bf Active sensing to characterize the heterogeneity of plant stress
}

\author{
Ayman Laaroussi$^1$ and
Peter Hanappe$^{1}$ and
David Colliaux$^{1}$ 
\thanks{This work was possible thanks to the support of the European Innovation Council Pathfinder Open DREAM (grant no. 101046451).}
\thanks{$^{1}$All authors are with \textit{Paris Research, Sony Computer Science Laboratories}, Paris, France
        {\tt\small ayman.laaroussi.91@gmail.com, firstname.surname@sony.com}}%
}




\maketitle

\begin{abstract}
While most phenotyping platforms rely primarily on image-based measurements, advanced plant characterization requires the integration of active physiological sensing modalities such as chlorophyll fluorescence. We present an autonomous robotic platform designed to perform targeted fluorescence measurements on plant leaves. The system combines 3D plant reconstruction, geometric analysis, and motion planning to localize suitable measurement points and generate collision-free trajectories for a robotic manipulator.

A dense 3D model of the plant is reconstructed from multi-view data and used to extract candidate leaf surfaces based on orientation, accessibility, and sensing constraints. These targets are then integrated into a task-level planning framework that guides the end-effector to precise contact or near-contact configurations required for point-based fluorescence acquisition. The platform enables automated, repeatable, and spatially resolved physiological measurements that go beyond passive imaging.

By tightly coupling perception, geometric reasoning, and manipulation, the proposed system provides a robotics-driven approach to high-resolution plant phenotyping and opens new directions for autonomous agricultural inspection and plant-aware manipulation.

\end{abstract}

\section{Introduction}

Plant phenotyping based on observable traits plays a critical role in modern agriculture, both for stress detection and for linking physiological responses to genotype in breeding programs. Vision-based systems can estimate macroscopic variables such as growth rate, biomass, or yield \cite{atefi2021robotic}. However, in the context of climate change, understanding how different plant varieties respond to environmental stress requires access to physiological indicators that go beyond passive imaging.

Chlorophyll fluorescence is a direct reporter of photosynthetic activity and provides a sensitive indicator of plant stress. Traditional fluorescence measurements in greenhouse or field conditions rely on handheld devices that require physically clamping individual leaves applying controlled light stimulation protocols \cite{gracia2025field}. While accurate, this approach is labor-intensive, requires manual intervention, and does not scale well to high-throughput phenotyping.

Recent advances in miniaturized electronics and the widespread availability of components developed for wearable devices have enabled the development of more compact fluorescence sensors. In this work, we employ such a portable device, kindly provided by the Jan IngenHousz Institute, which allows long-term, leaf-mounted recordings. Although these compact systems produce high-quality temporal data, their deployment still requires manual positioning and retrieval, making large-scale and spatially resolved measurements time-consuming and difficult to automate.


In this work, we present a robotic platform designed to automate targeted chlorophyll fluorescence measurements across multiple leaves and plants using a single portable sensor. The system integrates a robotic manipulator and a camera–sensor gimbal, enabling precise positioning of the fluorescence probe. We first describe the mechanical architecture of the platform and the integration and experimental validation of the fluorescence sensor. We then present our perception pipeline for leaf detection and 3D localization, followed by motion planning strategies that generate collision-free trajectories to position the sensor under geometric and accessibility constraints.

By coupling active physiological sensing with robotic perception and manipulation, the proposed system enables scalable, repeatable, and spatially resolved plant phenotyping, providing a robotics-driven solution for high-throughput agricultural inspection.




\section{Hardware}

We describe the hardware configuration of the active sensing platform. The system is designed to remain low-cost, with a total budget in the \$1000–1500 range, the majority of which corresponds to the CNC stage and the aluminum structural frame supporting the arm.

\subsection{The 5 Degrees-Of-Freedom Gimbal}

\begin{figure}
    \centering
    \includegraphics[width=0.45\linewidth]{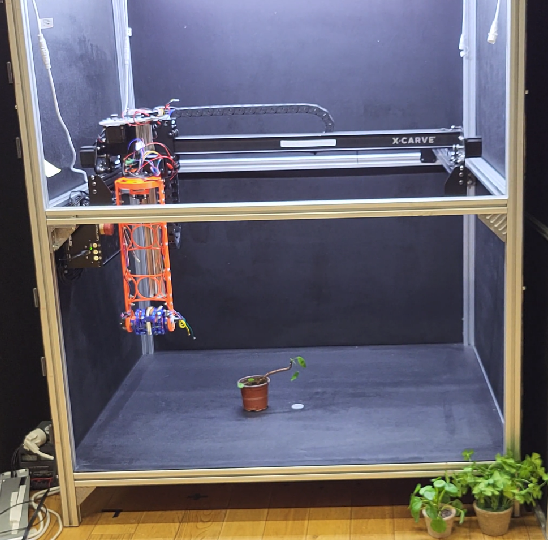}
    \includegraphics[width=0.483\linewidth]{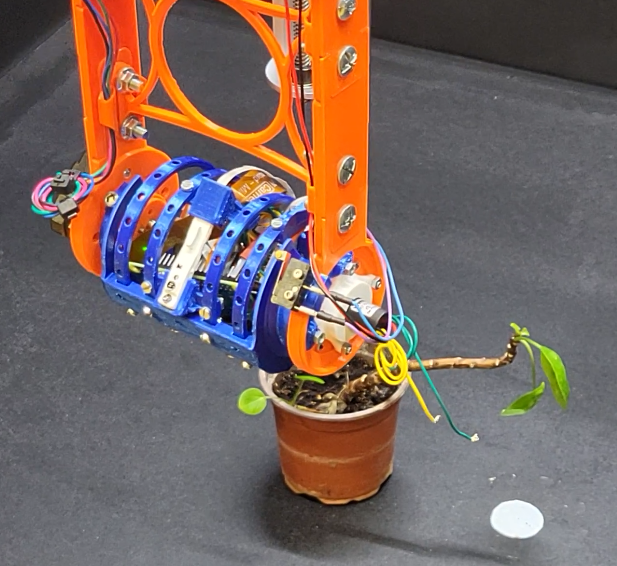}
    \caption{(Left) Plant scanner: A cartesian system mounted on an aluminium structure. (Right) Zoom on the gimbal with the sensor visible, the camera is on the opposite side.}
    \label{fig:robot}
\end{figure}

The sensing unit consists of a Pi Camera module 3 and an compact chlorophyll fluorescence sensor, mounted on a custom 3D-printed gimbal at the end-effector of a lightweight arm fabricated from PLA, PETG and aluminum elements (Fig. \ref{fig:robot}).

Cartesian positioning is achieved using an X-Carve CNC platform providing a 750×750mm planar workspace and 31.5 mm of z-axis motion, with a nominal accuracy of $\pm0.1mm$. Orientation control is provided by a two-axis pan–tilt mechanism driven by stepper motors, allowing full $360^\circ$ rotation in both axes. Two slip rings ensure continuous power delivery and signal transmission without cable torsion.

Motion control of the CNC axes is handled by motor drivers connected to a Raspberry Pi 5. The pan–tilt stepper motors are driven by an Adafruit Feather M0 interfaced via USB. The gimbal carries a Raspberry Pi Zero W for local acquisition and control of the camera and fluorescence sensor, communicating wirelessly with the main controller via Wi-Fi.

The system is organized into two distributed nodes: (i) the CNC motion node and (ii) the gimbal sensing node, communicating through RCOM\footnote{\url{https://github.com/romi/librcom}}, a custom middleware implementing a request–response protocol with automatic module discovery.




\subsection{The fluorescence sensor}

The device is an active chlorophyll fluorescence sensor. It operates by illuminating the leaf with an LED at the excitation wavelength of chlorophyll and measuring the emitted fluorescence signal, which serves as a direct proxy for photosynthetic activity. Accurate measurements require near-contact positioning (approximately 5 mm standoff distance) and alignment of the optical axis as close as possible to the local surface normal of the leaf. These geometric constraints motivate the need for precise robotic positioning and orientation control.

The sensor is interfaced via USB (through a serial adapter) to the gimbal sensing node, and communication is handled through a command-based serial protocol. The system allows programmable excitation sequences, enabling arbitrary light stimulation protocols depending on the experimental requirements. This flexibility makes the platform compatible with standard fluorescence assays as well as custom dynamic stimulation paradigms.

\begin{figure}
    \centering
    \includegraphics[width=.8\linewidth]{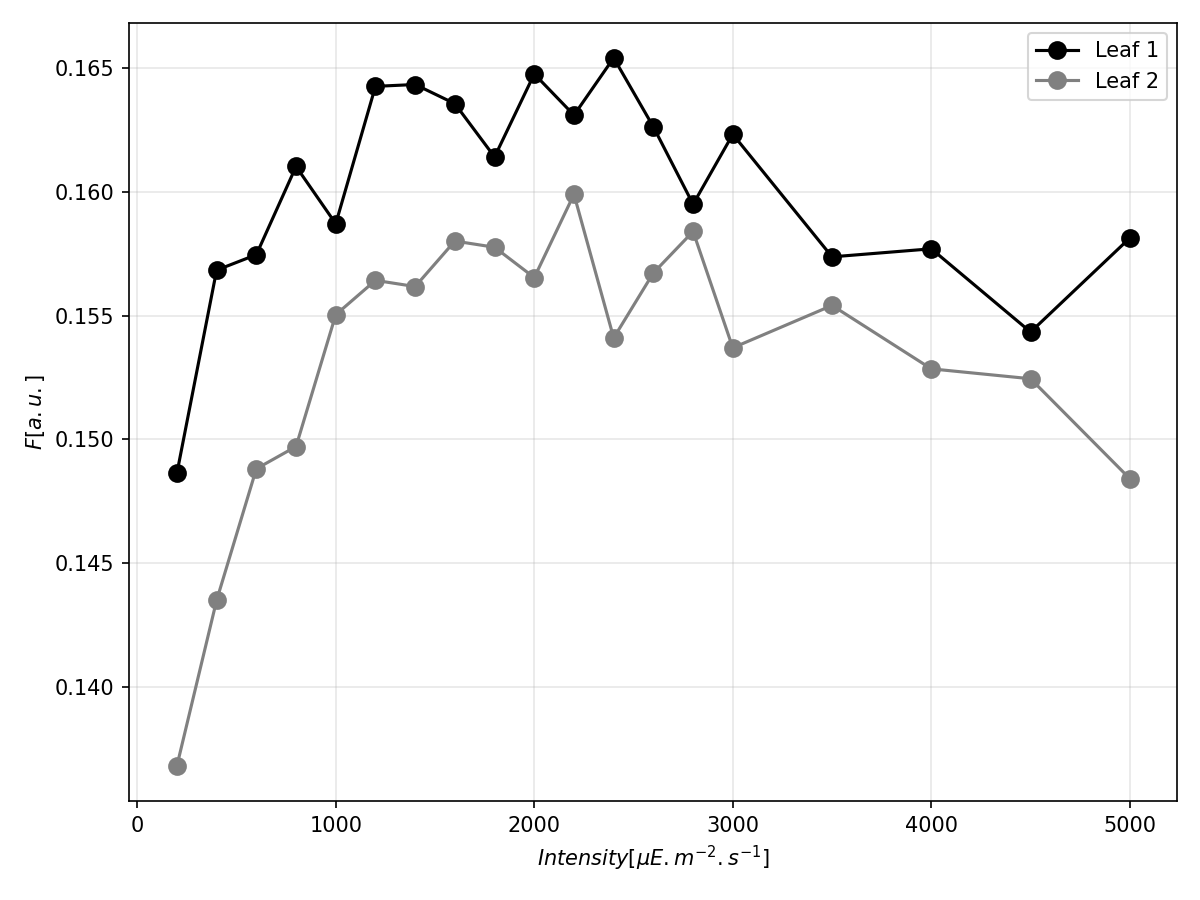}
    \includegraphics[width=\linewidth]{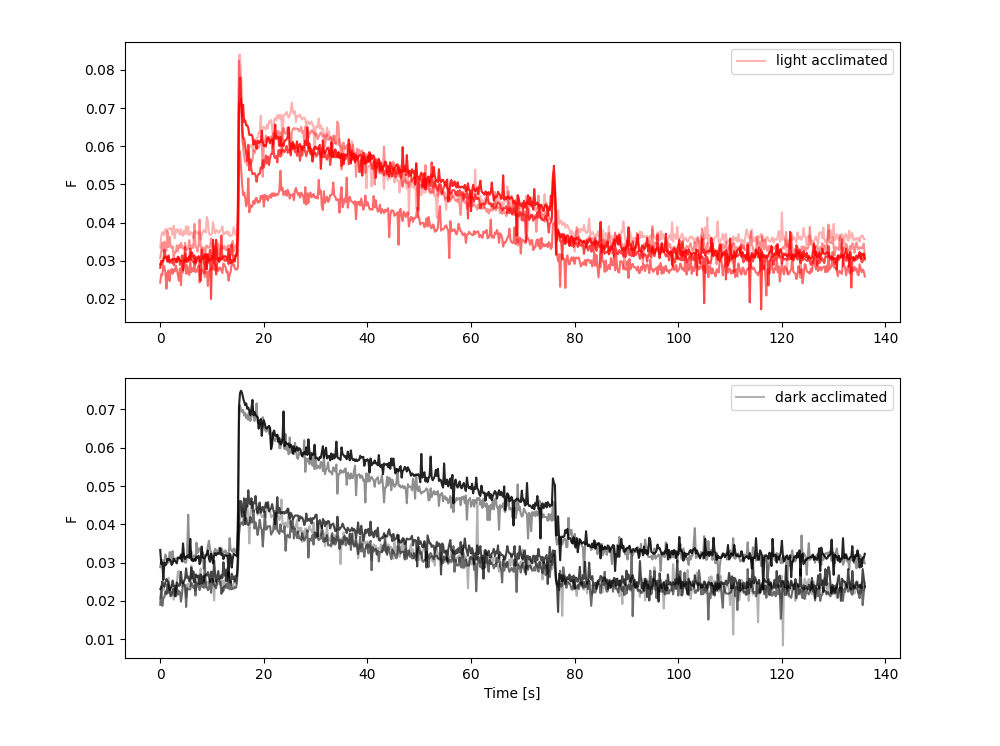}
    \caption{(Up) Chlorophyll fluorescence response depending on light intensity of the 1\ s pulse. (Bottom) Response to a sequence of Dark (25\ s) - Flash (1\ s) - Moderate light (1\ min) - Flash (1\ s) and Dark again (1min).}
    \label{fig:ambit}
\end{figure}

To characterize the sensor response, we applied a sequence of 1 s light pulses with increasing intensity, separated by 30 s dark intervals. The average fluorescence signal during each pulse was recorded (Fig. \ref{fig:ambit}) on two independent plants (Pilea peperomioides). The resulting light-response curves exhibit the expected behavior \cite{ralph2005rapid}: a linear increase in the low-light regime, followed by saturation and a gradual decrease at higher intensities due to the expression of a stress in response to the high-light, called non-photochemical quenching. These results confirm the sensor’s ability to capture physiologically meaningful fluorescence dynamics.

We further evaluated the sensor’s sensitivity to light-induced stress. One leaf was dark-acclimated, while a second leaf was exposed to high light for 30 min. Both leaves were then subjected to a stimulation protocol consisting of a 1 s saturating pulse ($5000\,\mu\mathrm{E}\,\mathrm{m}^{-2}\,\mathrm{s}^{-1}$, followed by 1 min of moderate light ($400\,\mu\mathrm{E}\,\mathrm{m}^{-2}\,\mathrm{s}^{-1}$), and a second 1 s saturating pulse. Fluorescence responses were recorded at five spatial locations per leaf. The two conditions produce markedly different fluorescence dynamics. In particular, the light-acclimated leaf exhibits a pronounced post-flash decrease followed by a relaxation phase characteristic of non-photochemical quenching processes.

From these measurements, we compute the ratio, derived from the relative fluorescence response during the transient decrease and the saturating flash. This scalar metric provides a compact and robust indicator of light stress, and can be readily integrated into automated robotic phenotyping workflows.

\section{Targeted measurements}

\begin{figure}
    \centering
    \includegraphics[width=0.98\linewidth]{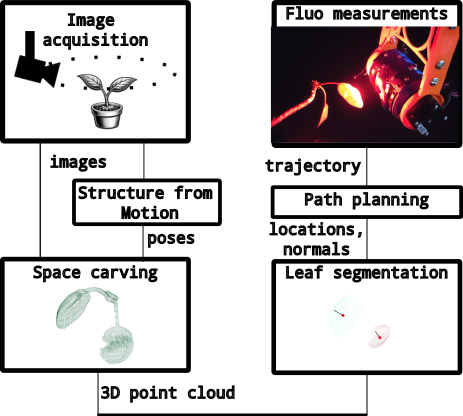}
    \caption{Summary of the perception and path planning pipeline.}
    \label{fig:pipeline}
\end{figure}

The fluorescence sensor requires precise positioning relative to the leaf. This is essential for reproducibility and controlling the level of noise in the signal. It is thus necessary to detect each leaf and set the pose of the sensor facing the surface plane of the leaf at the sensor working distance. The process for targeted measurements is summarized in Fig.\ref{fig:pipeline}.

\subsection{Perception component}

Leaf localization requires accurate 3D reconstruction of the plant followed by geometric segmentation of individual leaves. To reconstruct the plant geometry, we acquire multiple RGB images from viewpoints distributed in a circle around the specimen (80 views in our experiments). Camera poses are estimated using a Structure-from-Motion (SfM) pipeline implemented with COLMAP \cite{schonberger2016structure}.

Although SfM directly produces a sparse or dense point cloud, we instead use the recovered camera poses as input to a space carving algorithm \cite{wintz2018automated} to obtain a volumetric reconstruction. In our setup, space carving yields a denser and less noisy point cloud compared to the direct output of SfM. A known limitation, however, is that space carving assumes photo-consistency under silhouette constraints and cannot accurately reconstruct strong concavities. As a result, highly curled or self-occluding leaves may be partially reconstructed. Since the acquisition setup uses a uniformly black background, simple color-based segmentation is sufficient to isolate the plant from the scene. We compute an Excess Green (ExG) index to generate binary masks, which are then used as silhouettes for the 3D reconstruction pipeline.

From the reconstructed 3D point cloud, we perform leaf segmentation using local geometric descriptors. For each point, we compute the covariance matrix of its local neighborhood at multiple spatial scales (three scales corresponding to neighborhoods of approximately 30, 100, and 300 points). Eigenvalue decomposition of this covariance matrix provides a compact representation of local surface geometry.

Planar structures, such as leaf laminae, exhibit two dominant eigenvalues and one near-zero eigenvalue, while linear structures such as stems exhibit a single dominant eigenvalue. These geometric signatures allow for discrimination between leaf surfaces and supporting structures.

After classification, individual leaves are extracted using density-based clustering (HDBSCAN), which is robust to noise and variable point density. For each segmented leaf, we compute its centroid and estimate the local surface normal at that point. These geometric features are subsequently used to define target poses for the fluorescence sensor, ensuring orthogonal alignment and accessible positioning for robotic measurement.



\subsection{Motion planning}

Using the leaf centroids and surface normals estimated by the perception module, a motion planning strategy generates trajectories for the robotic arm that avoid collisions with the plant volume. The strategy operates in two phases: a global path planning step that routes the end-effector between successive target leaves while circumventing the plant, followed by a local approach step that positions the sensor at a geometrically consistent standoff distance from each leaf surface.

The plant volume is approximated as a vertical cylinder of radius $r_\text{cyl} = 0.15$\,m centred at the acquisition origin, spanning $z \in [-0.315, 0.0]$\,m. Before each displacement, the planner checks whether the straight line to the next waypoint intersects this volume by sampling 50 points along the segment. If an intersection is detected, intermediate waypoints are placed along a concentric
avoidance circle of radius $r_\text{avoid} = 0.25$\,m, distributed along the shorter arc between departure and arrival angles. Otherwise, the path reduces to a direct
connection with a single midpoint slightly offset in $z$ for smoothness. In both cases, the resulting control points are interpolated by a chord-length-parameterised cubic spline, sampled into 10--12 waypoints and transmitted as a single
\texttt{travel()} call to the CNC controller.

\begin{figure}
    \centering
    \includegraphics[width=0.98\linewidth]{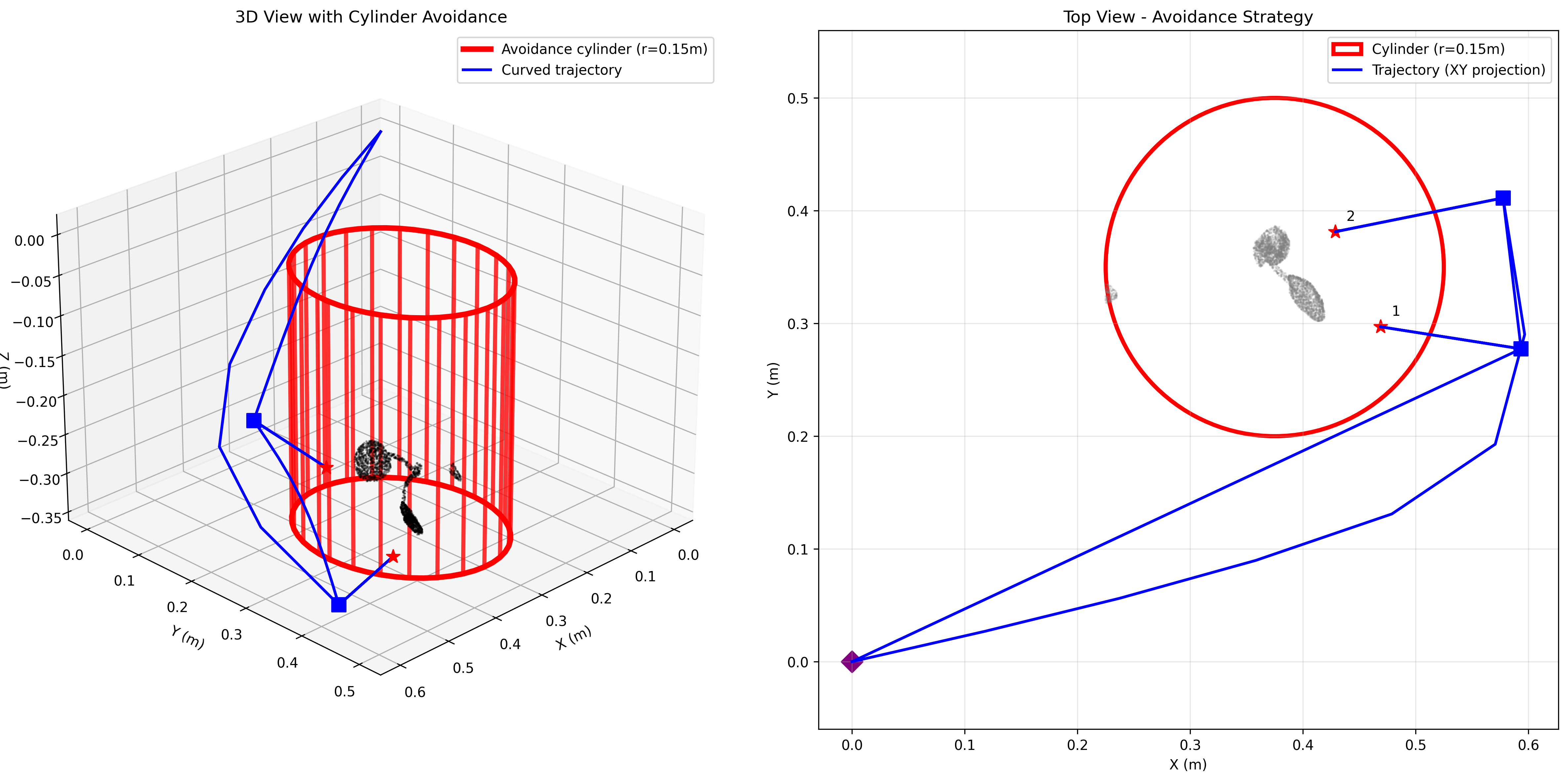}
    \caption{Illustration of the motion planning strategy showing global path planning with cylinder avoidance and local approach trajectories toward leaf targets at a $9$\,cm standoff distance.}
    \label{fig:path_planning}
\end{figure}

For each target leaf, the per-leaf acquisition sequence proceeds as follows. The end-effector first moves to an approach point defined as the leaf centroid displaced
by $d_\text{photo} = 9$\,cm along the outward surface normal. At this position, the gimbal orients the imaging sensor toward the centroid via analytically computed pan
and tilt increments, and an RGB image is acquired after a short stabilisation period. The gimbal then rotates $180°$ in tilt to bring the fluorescence sensor into alignment with the leaf. The camera and the fluorescence sensor are mounted back to back on the gimbal head, each displaced by $25$\,mm from the gimbal geometric centre, resulting in a total separation of $50$\,mm between the two sensor apertures.
To account for this geometry, the fluorescence target point is computed such that the gimbal geometric centre is positioned at $3$\,cm from the centroid along the outward
normal, with an additional correction of $25$\,mm applied along the vertical component projected onto the plane perpendicular to the leaf normal. This places the fluorescence sensor aperture within approximately $0$--$5$\,mm from the leaf surface for near-contact acquisition. After measurement, the arm returns to the approach point and the gimbal rotates $-180^\circ$ in tilt to restore its original orientation, after which the planner initiates the next spline segment toward the subsequent leaf.


\subsection{Visualization interface}

\begin{figure}
    \centering
    \includegraphics[width=0.98\linewidth]{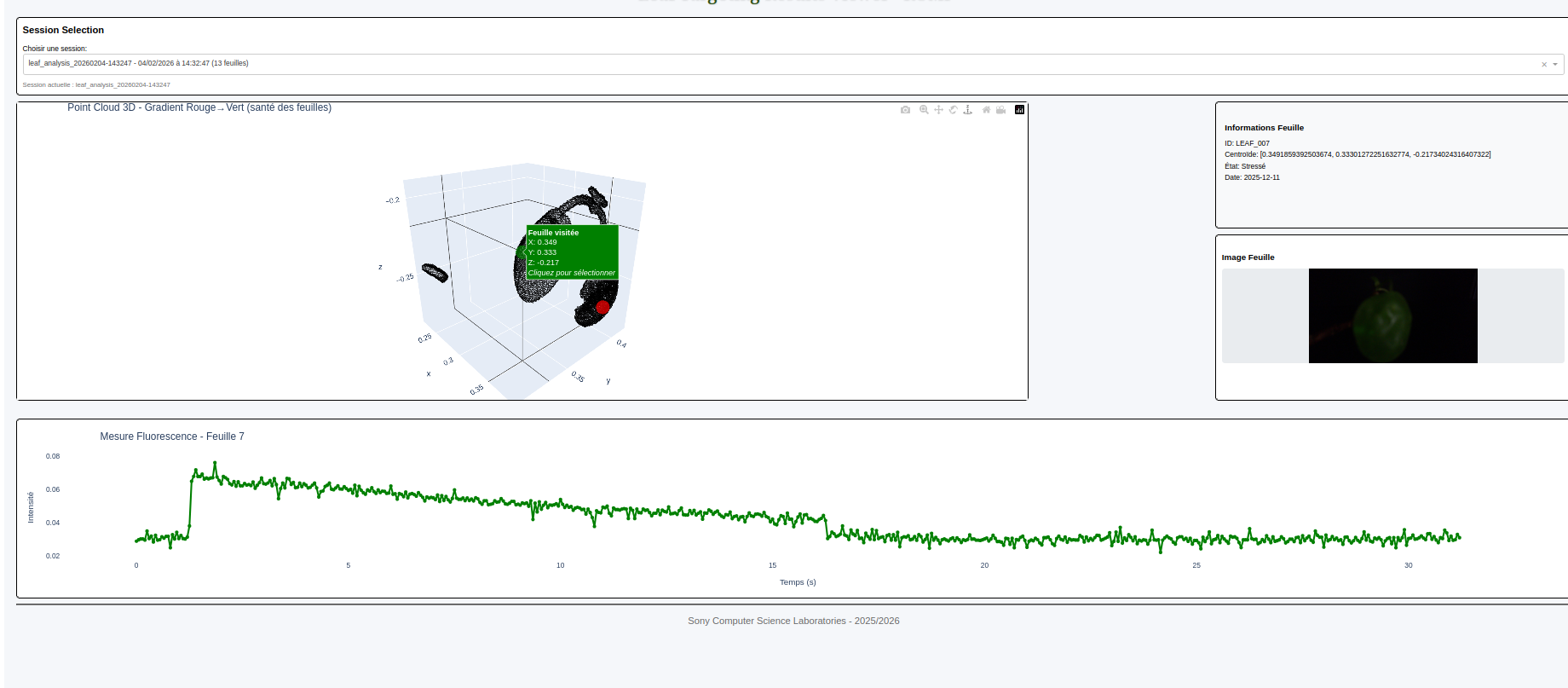}
    \caption{Web-based interface to inspect the fluorescence measurements: (Top-Left) Reconstructed point cloud with stress indicator reported in color. (Top-Right) Picture of the selected leaf. (Bottom) Recording from the fluorescence measurement.}
    \label{fig:interface}
\end{figure}

A web-based interface (Fig.\ref{fig:interface}) implemented using the Dash framework provides interactive visualization of the reconstructed point cloud and the corresponding measurement locations on each leaf. The interface allows users to inspect the spatial distribution of sampled points in 3D, alongside the associated RGB image of the probed leaf and the recorded fluorescence time series. This unified visualization enables qualitative assessment of targeting accuracy and physiological measurements.


\section{Conclusion}

We presented a low-cost robotic platform for automated active sensing of chlorophyll fluorescence in plants\footnote{The code is available at \url{https://github.com/SonyCSLParis/Plant3DImager}}. By integrating 3D perception, geometric reasoning, and precise manipulation, the system enables targeted, repeatable fluorescence measurements at the leaf level. Using the proposed stimulation protocol, we demonstrated that the platform can discriminate between dark-adapted and light-exposed leaves based standard indicator of photosynthetic light stress.

While the binary discrimination between dark-adapted and light-stressed leaves validates the sensing and positioning capabilities of the system, more subtle stress heterogeneity will require higher-dimensional analysis of fluorescence dynamics. In this context, machine learning approaches such as dictionary learning \cite{lahlou2025interplay} could be leveraged to extract informative features from temporal fluorescence traces and improve sensitivity to complex physiological states. The results could also allow guiding the placement of leaf-mounted devices based on heterogeneity across the plant.

The current evaluation was conducted on plants with relatively simple geometry. For specimens with highly complex or self-occluding structures, the reconstruction and segmentation pipeline may fail due to the limitations of the space-carving approach. More advanced 3D reconstruction methods, such as recent neural methods (e.g., Mast3r \cite{leroy2024grounding}), could handle more complex geometries, potentially at the cost of noisier point clouds.

Finally, reducing reconstruction time is critical for deployment in greenhouse or field conditions. Future work will focus on accelerating the perception pipeline (eg. by reducing the number of views) and we are also interested in the integration of the sensing arm onto a mobile robotic platform, such as the one developed in the ROMI project\footnote{See for example at \url{https://www.youtube.com/watch?v=4XBq29rmo5E}}, enabling scalable and autonomous plant-level phenotyping in real agricultural environments.




\section*{Acknowledgment}
The authors would like to thank Ludovico Caracciolo and David Kramer from the Jan IngenHousz Institute who allowed us to use a preliminary version of their fluorescence instrument and Aliénor Lahlou from Paris Research, Sony CSL for fruitful discussions and her help in calibration of the instrument. An LLM assistant (ChatGPT 5.2) was used for clarifiying and correcting language in the writing of this paper. All scientific content and interpretations are solely the responsibility of the authors.

\bibliographystyle{IEEEtran}
\bibliography{references}
\end{document}